\documentclass[times,review,10pt]{elsarticle}

\usepackage{amssymb}
\usepackage{amsmath}
\usepackage{algorithm}
\usepackage{multirow}
\usepackage{booktabs}
\usepackage{bbding}
\usepackage{hyperref}
\hypersetup{hypertex=true,
    colorlinks=true,
    linkcolor=ccr,
    anchorcolor=ccr,
    citecolor=ccr}

\journal{Pattern Recognition}

\begin{document}

\begin{frontmatter}

\title{IDEA: Image Description Enhanced CLIP-Adapter}

\author[aff1]{Zhipeng Ye} 
\ead{zhipengye@nustti.edu.cn}
\author[aff1]{Feng Jiang \corref{cor1}}
\ead{jf@nustti.edu.cn}
\author[aff2]{Qiufeng Wang}
\ead{qiufeng.wang@xjtlu.edu.cn}
\author[aff3]{Kaizhu Huang}
\ead{kaizhu.huang@dukekunshan.edu.cn}
\author[aff1]{Jiaqi Huang}
\ead{2207880112@nustti.edu.cn}

\affiliation[aff1]{organization={Taizhou Institute of Science and Technology, Nanjing University of Science and Technology},
            city={Taizhou},
            postcode={225300}, 
            state={Jiangsu},
            country={China}}
\affiliation[aff2]{organization={Department of Intelligence Science, Xi’an Jiaotong-Liverpool University},
            city={Suzhou},
            postcode={215123}, 
            state={Jiangsu},
            country={China}}
\affiliation[aff3]{organization={Duke Kunshan University},
            city={Suzhou},
            postcode={215123}, 
            state={Jiangsu},
            country={China}}
            
\cortext[cor1]{Corresponding author}

\begin{abstract}
CLIP (Contrastive Language-Image Pre-training) has attained great success in pattern recognition and computer vision. Transferring CLIP to downstream tasks (e.g. zero- or few-shot classification) is a hot topic in multimodal learning. However, current studies primarily focus on either prompt learning for text or adapter tuning for vision, without fully exploiting the complementary information and correlations among image-text pairs. In this paper, we propose an Image Description Enhanced CLIP-Adapter (IDEA) method to adapt CLIP to few-shot image classification tasks. This method captures fine-grained features by leveraging both visual features and textual descriptions of images. IDEA is a training-free method for CLIP, and it can be comparable to or even exceeds state-of-the-art models on multiple tasks. Furthermore, we introduce Trainable-IDEA (T-IDEA), which extends IDEA by adding two lightweight learnable components (i.e., a projector and a learnable latent space), further enhancing the model's performance and achieving SOTA results on 11 datasets. As one important contribution, we employ the Llama model and design a comprehensive pipeline to generate textual descriptions for images of 11 datasets, resulting in a total of 1,637,795 image-text pairs, named "IMD-11". Our code and data are released at \href{https://github.com/FourierAI/IDEA}{https://github.com/FourierAI/IDEA}.

\end{abstract}

\begin{highlights}
\item A new training-free adapter termed as IDEA is proposed for CLIP.
\item As a first attempt, IDEA explores complement and correlation information among image-text pairs.
\item IDEA achieves state-of-the-art performance in extensive experiments.
\item A new multimodal dataset is released to the community.
\end{highlights}


\end{frontmatter}



\section{Introduction}
\label{sec:intro}

While animals primarily perceive the world through their visual system, only humans have evolved language systems over millions of years. Language enables humans to understand, use, and create things in a logical reasoning manner, ultimately evolving into intelligence. In computer vision, some studies~\cite{clip,ALIGN,ALIP} have shown that incorporating language/text information into vision tasks can significantly enhance a model's visual understanding capabilities and therefore improve its performance. CLIP (Contrastive Language-Image Pre-training)~\cite{clip} is a dual-tower structure Vision-Language Model (VLM) that consists of a visual encoder and a textual encoder. CLIP is pre-trained on large-scale image-text pairs using contrastive learning. During this process, the image and text data interact with each other, endowing the model with a generalization ability and leading CLIP to be able to classify unseen images during training (called zero-shot learning)\cite{3dzs,bzs}.

\begin{figure}[htb!]
    \centering
    \includegraphics[width=0.99\linewidth]{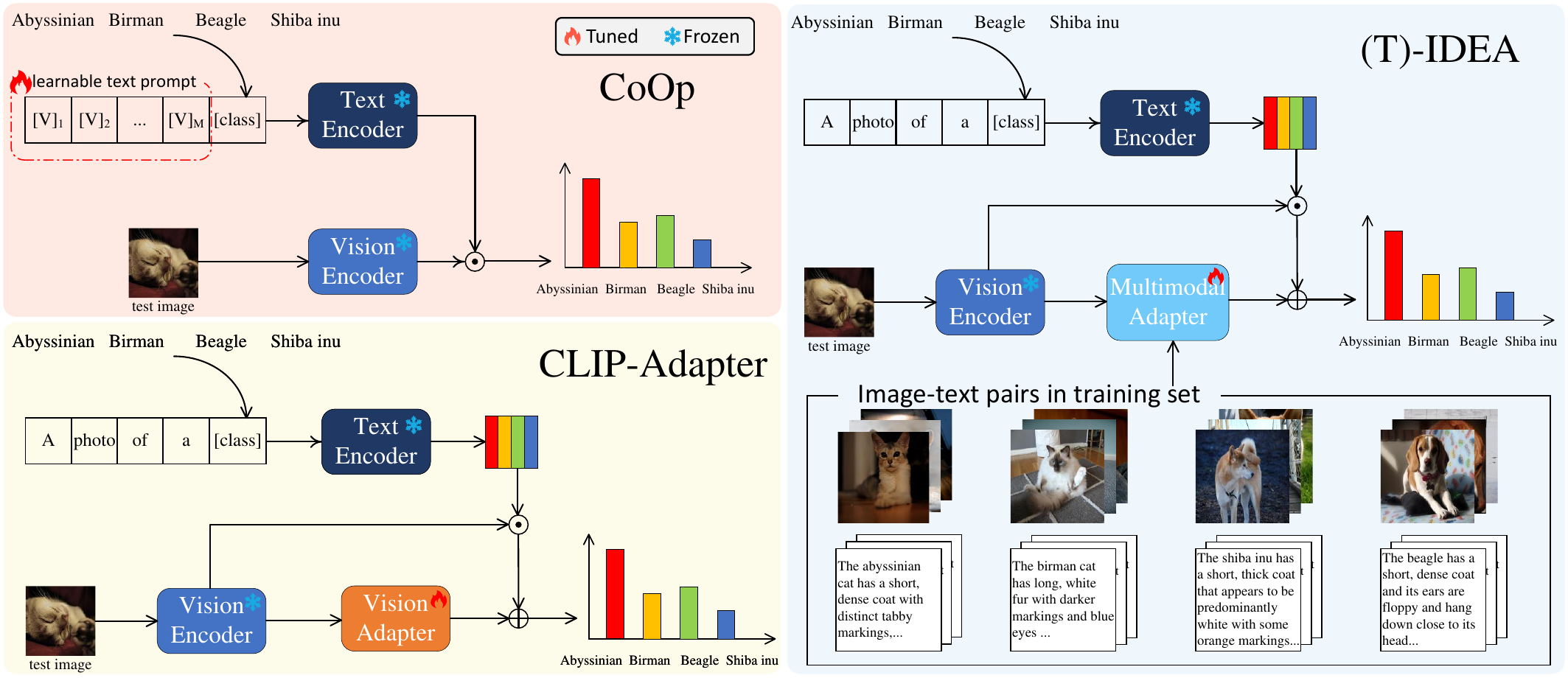}
    \caption{Comparison of different PEFTs for CLIP. CoOp~\cite{coop} and CoCoOp~\cite{cocoop} have a similar architecture. Tip-Adapter~\cite{tip} shares the same architecture as CLIP-Adapter~\cite{clip-adapter}. Different from previous works, (T)-IDEA introduces a multimodal adapter that explores the complementary relationship and semantic correlation among image-text pairs.}
    \label{fig:comparision}
\end{figure}

Fine-tuning CLIP for downstream vision tasks has become a hot research topic in recent years~\cite{clip-art,PLIP}. Notably, PEFT is a novel fine-tuning method, which freezes the parameters of the model’s backbone and fine-tunes the incorporated lightweight learnable parameters on downstream datasets~\cite{PEFT}. PEFT achieves or even surpasses the performance of full fine-tuning on some tasks. Recent studies focus on exploring PEFT  for CLIP. Linear Probe~\cite{clip} utilizes CLIP's vision encoder to extract features, which are subsequently fed to a linear layer for training, enabling it to handle few-shot image classification tasks where a very limited number of samples are available for each class of data \cite{prfs,prcvfs,qqfs}. Subsequent research~\cite{coop,cocoop} focuses on exploiting text features to enhance the performance of few-shot learning. As shown in Fig.~\ref{fig:comparision}, CoOp~\cite{coop} and CoCoOp~\cite{cocoop} improve few-shot image classification performance by incorporating learnable text prompts. CLIP-Adapter~\cite{clip-adapter} optimizes a vision adapter, which is a two-layer Multi-Layer Perceptron (MLP), to learn new vision features for few-shot image classification tasks. In Training-Free CLIP-Adapter (Tip-Adapter)~\cite{tip}, the two-layer MLP is replaced by a cache model, leading to a significant boost in the performance of few-shot image classification tasks.

The work mentioned above primarily concentrates on optimizing text prompts or vision adapters, without fully exploiting the complementary relationship and inherent semantic correlation among image-text pairs, thus limiting their performance. To this end, in this paper, we propose a novel multimodal adapter, Image Description Enhanced CLIP-Adapter (IDEA), where the test image is retrieved against image-text pairs from the training set for few-shot classification tasks. IDEA is a training-free method, yet its performance rivals that of supervised training methods. Furthermore, we introduce Trainable Image Description Enhanced CLIP-Adapter (T-IDEA), which adopts two learnable components in IDEA to promote the model's performance further. T-IDEA achieves state-of-the-art (SOTA) performance on 11 commonly used image datasets. In addition, most of the image datasets lack corresponding image descriptions, and labeling these datasets is time-consuming and laborious. We employ Llama~\cite{llama} and design a comprehensive pipeline to generate a textual description for each image. 

The contributions of this paper are summarized as follows:
\begin{itemize}
    \item We propose IDEA, which utilizes the complementary relationship among image-text pairs and explores semantic association across multi-modalities in a training-free paradigm, to realize few-shot image classification. 
    
    \item We propose T-IDEA, which extends IDEA by adopting a lightweight projection layer and a learnable semantic latent space to boost the performance of IDEA.
    
    \item We design a comprehensive pipeline to generate image descriptions for 11 public image datasets, resulting in a total of 1,637,795 image-text pairs. Our dataset, which is referred to as "IMD-11", has been made publicly available.
     
    \item We evaluate the proposed methods on 11 public image datasets. The experimental results show that IDEA and T-IDEA respectively outperform SOTA methods in the training-free and training-required settings.
    
\end{itemize}

\section{Related Work}
\label{sec:related_work}
In this section, we review the literature related to the paper, including Vision-Language Model (VLM) and Parameter-Efficient-Fine-Tuning (PEFT).

\subsection{Vision-Language Model}
Modality is the way humans perceive and recognize the world, which includes vision, text, auditory, touch, and etc \cite{vl_survey, m_survey, vl_survey2}. For humans, vision and text are the main ways they perceive the world, which has attracted extensive research interests~\cite{PLIP, RCLIP, vl-fmm} from scholars around the world. The invention of Transformer~\cite{atis} provides a unified model for both computer vision (CV) and natural language processing (NLP), and gives birth to the development of VLM~\cite{clip}. VLM is a kind of pre-training model, the training methods of which are mainly categorized into Image-Text Contrastive Learning and Pre-training with Generative Objectives~\cite{vl_survey}.

\textbf{Image-Text Contrastive Learning.} Image-Text Contractive Learning is the most common method for training VLMs. It employs contrastive learning to process input image-text pairs, ensuring that the image-text pairs with similar semantics are close in the embedding space, and the image-text pairs with different semantics are far away in the embedding space. CLIP~\cite{clip} collects and cleans 400 million image-text pairs from the internet, and pre-trains them by using InfoNCE loss. Subsequently, CLIP performs zero-shot image classification by evaluating similarities between test samples and category names. The same pre-training method is adopted in ALIGN~\cite{ALIGN}, which collects 1.8 billion noisy image-text pairs and gets a favourable result as well. The success of ALIGN verifies that multimodal could learn good Vision-Language representations by enlarging the size of training data, even if there might be massive noises in the data. ALIP~\cite{ALIP} assumes that the image collected from the internet is noisy and generates a caption for each of them by Large Language Model (LLM), after which a dual-path model is pre-trained on the generated captions and the raw texts. ZeroVL~\cite{zerovl} proposes a comprehensive pipeline, i.e. Debiased Data Sampling and Coin Flipping Mixup, to train the model with a limited training set of 14 million samples. PyramidCLIP~\cite{PyramidCLIP} achieves fine-grained semantic alignment through cross-level and peer-level contrastive learning.

\textbf{Pre-training with Generative Objectives.} Pre-training with Generative Objectives is the other major method for training VLM, it masks parts of the raw data and regenerates the masked content by context, and thus learns the semantic correlations among various modalities. KELIP~\cite{KELIP} splits an image into several patches and randomly masks some of them, then recovers the masked patches by image context which is also used in MAE~\cite{He2021MaskedAA}. SegCLIP~\cite{segclip} proposes a reconstruction loss and a superpixel-based KL loss to enhance the model’s visual representation, achieving open-vocabulary image segmentation in an annotation-free manner. FIBER~\cite{FIBER} integrates contrastive loss, generative loss, and alignment loss to propose a deep multimodal fusion method for coarse-to-fine pre-training of the VLM. FLAVA~\cite{FLAVA} masks 40 percent image patches and 15 percent text tokens, and subsequently predicts the masked patches and tokens with the use of MLP, to better capture the correlation between vision and language. The above-mentioned methods pre-train the VLM by recovering parts of images or texts, some other models could even recover the full image or image description from image-text pairs. Diffusion models~\cite{dm1,dm2,dm3} use text as the prompt to generate the corresponding image that is consistent with the text. COCA~\cite{coca} adopts an encoder-decoder architecture to pre-train the VLM, where the input is the image, and the output is the caption corresponding to the image. LLaVA~\cite{llava} uses the pre-trained CLIP image encoder to obtain the image features and then converts the image features into text tokens through a trainable projection layer, which could remarkably promote the multimodal understanding ability of the model. Llama~\cite{llama} employs a lightweight projection layer to align the input image with text and performs inter-modal fusion with the use of a cross-attention mechanism.

\subsection{Parameter-Efficient Fine-Tuning}

In the past few years, fine-tuning the pre-trained models on large-scale datasets to adapt downstream tasks has dominated the deep learning paradigm. However, there are significant disadvantages with this approach~\cite{PEFT}. Firstly, in terms of large-scale models, full fine-tuning is challenging, time-consuming, and unsustainable. Secondly, fine-tuning large models on downstream tasks could potentially cause catastrophic forgetting. To tackle these issues, some scholars proposed Parameter-Efficient Fine-Tuning (PEFT)~\cite{PEFT,peft2}. PEFT is a new fine-tuning method, it freezes all the parameters of backbone and adapts the model to different downstream tasks by fine-tuning the parameters of additional modules attached to the model. PEFT is roughly categorized into two types: Prompt Tuning, and Adapter Tuning.

\textbf{Prompt tuning.} Prompt tuning adapts downstream tasks by adding learnable tokens in either input or hidden layers of the model as learnable prompts. Inspired by prompt learning in NLP, VPT~\cite{vpt} adopts prompt tuning technology to vision tasks for the first time. VPT fine-tunes the model by adding some prompt tokens in the input space and outperforms most full fine-tuning methods in multiple tasks. CoOp~\cite{coop} employs learnable token vectors instead of manually designed prompts as input for the text encoder, achieving commendable performance in few-shot image classification tasks. On the basis of CoOp, CoCoOp~\cite{cocoop} designs a lightweight neural network to generate prompts for each image, which is known as Conditional Prompt Learning.

\textbf{Adapter tuning.} In adapter tuning, the model is equipped with additional learnable layers (e.g. MLP, Transformer~\cite{transformer}) 
to adapt to the downstream tasks. CLIP-Adapter~\cite{clip-adapter} uses an additional lightweight bottleneck layer (i.e. two linear layers following the last layer of vision encoder and text encoder, respectively) to learn new features and fuses them with the original pre-trained features via residual connection. Tip-Adapter~\cite{tip} makes use of key-value pairs collected from the few-shot training set to construct the adapter, which is called cache model. The linear layers in the CLIP-Adapter is replaced by the cache model, rendering Tip-Adapter a training-free method and being superior to other few-shot classification methods. Based on the Tip-Adapter, the keys in the cache model are dynamically updated using stochastic gradient descent, further enhancing the performance of Tip-Adapter and achieving the SOTA result.

\section{Method}
\label{sec:method}

In this section, we elaborate on the proposed methods. Firstly, we briefly review the zero-shot image classification of CLIP. Then, we describe IDEA and T-IDEA in detail, respectively. Last, we introduce the process of generating image descriptions. 

\subsection{Revisiting zero-shot image classification of CLIP}
The CLIP model is trained on a large-scale image-text pairs dataset by contrastive learning. It mines the semantic association between image-text pairs and enables the model to obtain a high generalization ability, achieving SOTA results in several vision downstream tasks. CLIP adopts a zero-shot classification strategy where the test image is retrieved against the textual information of category names to find the most matching category as the classification result. This ensures that CLIP can achieve open-vocabulary classification without re-training.

Specifically, given a test image $\mathbf{I}_{\text{test}}$, we feed it into the vision encoder of CLIP to obtain the corresponding visual feature $i_{\text{test}}\in \mathbb{R}^{D\times 1}$, where $D$ is the dimension of the visual feature. Eq.~\ref{eq:testenc} describes the process.
\begin{equation}\label{eq:testenc}
    i_{\text{test}} = \text{VisionEncoder}(\mathbf{I}_{\text{test}})
\end{equation}

Then, let $N$ be the number of categories and $S_{\text{label}}$ be the set of category names. A manually designed prompt template (e.g. a photo of a \{object\}) is used to generate a textual prompt template for each of the categories. Next, the textual prompts are fed into the CLIP's text encoder to obtain the corresponding features $\mathbf{T}_{\text{class}} \in \mathbb{R}^{N\times D}$, as shown in Eq.~\ref{eq:promenc}.
\begin{equation}\label{eq:promenc}
    \mathbf{T}_{\text{class}} = \text{TextEncoder}(\text{Template}(S_{\text{label}}))
\end{equation}

Finally, we get the output $\text{logits} \in \mathbb{R}^{N\times 1}$ for classification, as denoted in Eq.~\ref{eq:clip-zero}.
\begin{equation}\label{eq:clip-zero}
    \text{logits} = \underbrace{\mathbf{T}_{\text{class}} \cdot  i_{\text{test}}}_{\text{zero-shot knowledge}}
\end{equation}
where $\cdot$ means matrix multiplication, and both $\mathbf{T}_{\text{class}}$ and $i_{\text{test}}$ are normalized in the feature dimension.
The classification result of CLIP is the index corresponding to the maximum value of the logits. For convenience, we refer to $\mathbf{T}_{\text{class}} \cdot  i_{\text{test}}$ in Eq.~\ref{eq:clip-zero} as the zero-shot knowledge.

\subsection{Image Description Enhanced CLIP-Adapter}

\begin{figure}[htb!]
    \centering
    \includegraphics[width=0.99\linewidth]{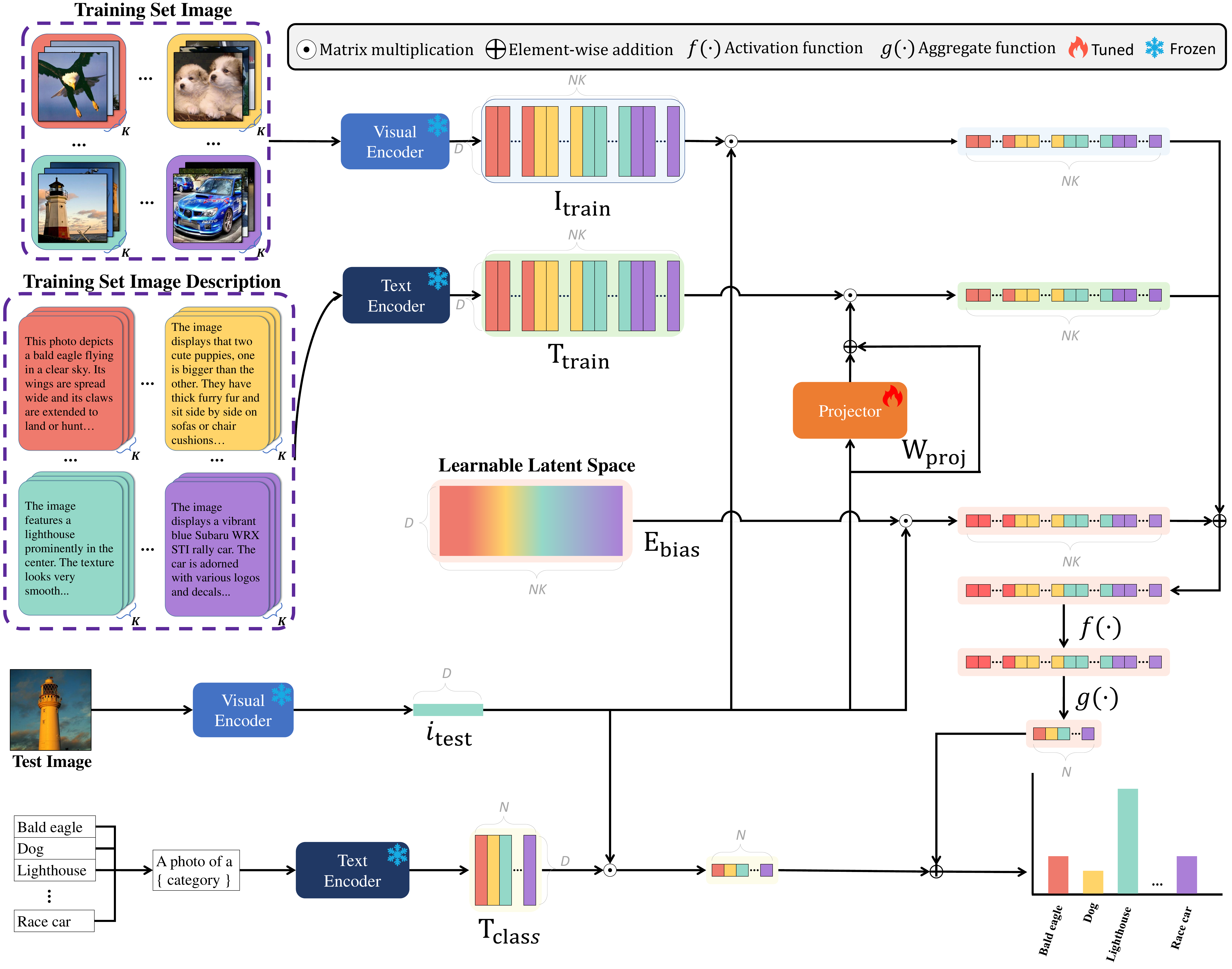}
    \caption{The architecture of IDEA and T-IDEA. Given a training set with $K$-shot and $N$-class, CLIP encodes visual and textual data to obtain $\mathbf{I}_{\text{train}}$ and $\mathbf{T}_{\text{train}}$, respectively. Then, we compute and convert the instance-level similarity into class-level similarity as few-shot knowledge. Additionally, we design a trainable projector $\mathbf{W}_{\text{proj}}$ and a learnable latent space $\mathbf{E}_{\text{bias}}$ to improve performance. Finally, we combine the few-shot knowledge with the original zero-shot knowledge to get the model logits.}
    \label{fig:framework}
\end{figure}

Based on the zero-shot classification with CLIP, we propose an novel adapter called  Image Description Enhanced CLIP-Adapter (IDEA), in which we explore the few-shot knowledge from the image-text pairs to strengthen CLIP.

Firstly, we construct a $K$-shot $N$-class training set that contains both visual information and textual descriptions of the images. Then, we freeze the parameters of CLIP's visual and textual encoders for PEFT. Eq.~\ref{eq:supenc} demonstrates that the $\mathsf{image}$ in the training set is fed into the visual encoder to get the visual features $\mathbf{I}_{\text{train}}\in \mathbb{R}^{NK\times D}$, and the $\mathsf{text}$ in the training set is fed into the textual encoder to get the textual features $\mathbf{T}_{\text{train}} \in \mathbb{R}^{NK\times D}$.

\begin{equation}\label{eq:supenc}    
\begin{aligned}
    &\mathbf{I}_{\text{train}} = \text{VisionEncoder}(\mathsf{Image})\\
    &\mathbf{T}_{\text{train}} = \text{TextEncoder}(\mathsf{Text})
\end{aligned}
\end{equation}

Subsequently, we compute the multimodal similarities, as shown in Eq.~\ref{eq:sim_idea}.
\begin{equation}
\begin{aligned}\label{eq:sim_idea}
    &Sim_{I} = \mathbf{I}_{\text{train}} \cdot i_{\text{test} }\\
    &Sim_{T} = \mathbf{T}_{\text{train}} \cdot i_{\text{test}} 
\end{aligned}
\end{equation}
where $Sim_{I} \in \mathbb{R}^{NK\times 1}$ is the similarity between the test image and the images in the training set, $Sim_{T} \in \mathbb{R}^{NK\times 1}$ is the similarity between the test image and the textual description in the training set. 

IDEA computes the similarity between the test image and $K$ samples of each category in the training set, which is referred to as few-shot knowledge and facilitates the mining of fine-grained semantic correlations between images and texts. Previous studies~\cite{clip,coop} indicate that incorporating textual information into visual models can effectively enhance their logical reasoning capabilities. Thus, we utilize both visual and textual information in the training set to promote recognition ability.

\begin{figure}[ht!]
    \centering
    \includegraphics[width=0.99\linewidth]{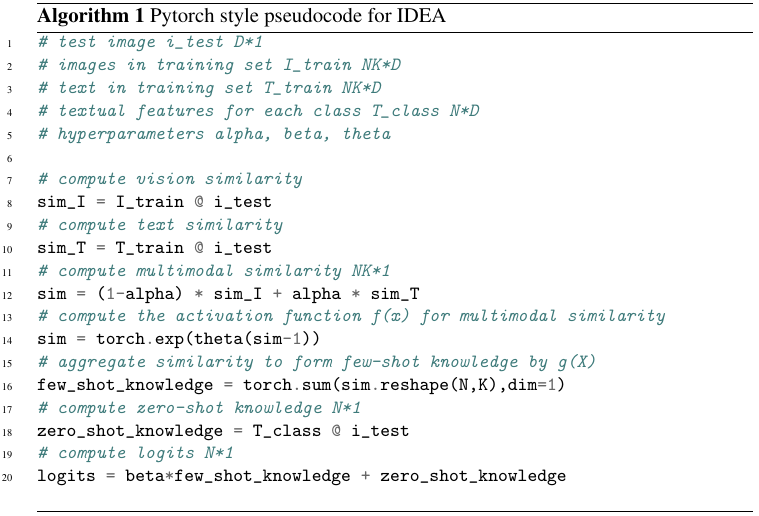}
    \caption{Pytorch style pseudocode for IDEA}
    \label{fig:algorithm}
\end{figure}

Finally, by combining zero-shot knowledge and few-shot knowledge, we get the output $\text{logits} \in \mathbb{R}^{N\times 1}$, as shown in Eq.~\ref{eq:log_idea}:
\begin{equation}\label{eq:log_idea}
\begin{aligned}
    \text{logits} &= \beta \underbrace{ g\{ f[ (1-\alpha) Sim_I + \alpha Sim_T) ]\} }_{\text{Few-Shot Knowledge}} \\
    &+ \underbrace{\mathbf{T}_{\text{class}} \cdot i_{\text{test}}}_{\text{Zero-Shot Knowledge}}
\end{aligned}
\end{equation}
where $\alpha \in [0,1]$ is a hyperparameter to balance the similarity between visual modal and textual modal, and $\beta \in (0, \infty)$ is a hyperparameter to trade off the few-shot knowledge and the zero-shot knowledge. The activation function $f(x) = \text{exp}(\theta(x-1))$ is defined for mapping the value of similarity to the interval [0,1]. $\theta \in (0, \infty)$ controls the sharpness of the activation function, which dynamically stretches and compresses the value of similarity to better fuse the few-shot knowledge to the zero-shot knowledge. Given the multimodal similarity among samples $\mathbf{X} \in \mathbb{R}^{NK\times 1}$, we define a function $g(\mathbf{X}) = \sum_{K} \text{reshape}(\mathbf{X}, N, K)$ to aggregate the similarity among samples to form the few-shot knowledge. $g(\mathbf{X})$ reshapes $\mathbf{X}$ into a matrix with $N$ rows and $K$ columns. We then sum the matrix in the column dimension. This operation is used to aggregate the instance-level similarity into the class-level similarity.
Fig.~\ref{fig:algorithm} shows pseudocode of the core of an
implementation of IDEA.

The advantages of IDEA are summarized as follows. Firstly, IDEA utilizes the corresponding textual descriptions of the images as a supplement to visual information, which improves the performance of the CLIP's few-shot image classification. Secondly, IDEA combines the knowledge of zero-shot and few-shot to capture fine-grained semantic correlations of image-text pairs, which enhances the fusion of multimodal data. Finally, IDEA is a training-free method for CLIP which can be comparable to or even outperform the SOTA models.

\subsection{Trainable Image Description Enhanced CLIP-Adapter}

IDEA does not require stochastic gradient descent (SGD) to train the model and exhibits strong recognition performance in few-shot classification tasks. Even so, we believe that the performance of IDEA can be further improved. Therefore, we propose a Trainable Image Description Enhanced CLIP-Adapter (T-IDEA) method.

On one hand, we believe that there is a native intermodal semantic gap between visual and textual information when calculating the item of $\mathbf{T}_{\text{train}}\cdot i_{\text{test}}$ in Eq.~\ref{eq:sim_idea}. To overcome this problem, we design a lightweight projection layer $\mathbf{W}_{\text{proj}}\in \mathbb{R}^{D\times D}$ for intermodal semantic alignment, and utilize residual connection for modal fusion.

On the other hand, for few-shot image classification tasks, the selected $K$ samples cannot supplemently cover all the samples in the training set, which means there is a bias of semantics between the $K$ samples and all the samples. Therefore, we design a trainable semantic latent space $\mathbf{E}_{\text{bias}}\in \mathbb{R}^{NK\times D}$ to correct the bias in the semantic space of the training set.

Therefore, the formula for the logits of the T-IDEA is defined in Eq.~\ref{eq:T-IDEA}.
\begin{equation}
    \begin{aligned}
        \text{logits} &= \beta g \{ f[ (1-\alpha) \mathbf{I}_{\text{train}} \cdot i_{\text{test}} + \alpha( \underbrace{\mathbf{T}_{\text{train}} \cdot \mathbf{W}_{\text{proj}} \cdot i_{\text{test}} + \mathbf{T}_{\text{train}} \cdot i_{\text{test}})}_{\text{align intermodal semantic}} \\ 
        &+ \underbrace{\mathbf{E}_{\text{bias}} \cdot i_{\text{test}}}_{\text{correct semantic bias}} ] \} + \mathbf{T}_{\text{class}} \cdot i_{\text{test}} 
    \end{aligned}
    \label{eq:T-IDEA}
\end{equation}
where $\mathbf{W}_{\text{proj}}\in \mathbb{R}^{D\times D}$ and $\mathbf{E}_{\text{bias}}\in \mathbb{R}^{NK\times D}$ are lightweight components.

\subsection{Image Description Generation}

\begin{figure}[ht!]
    \centering
    \includegraphics[width=0.99\linewidth]{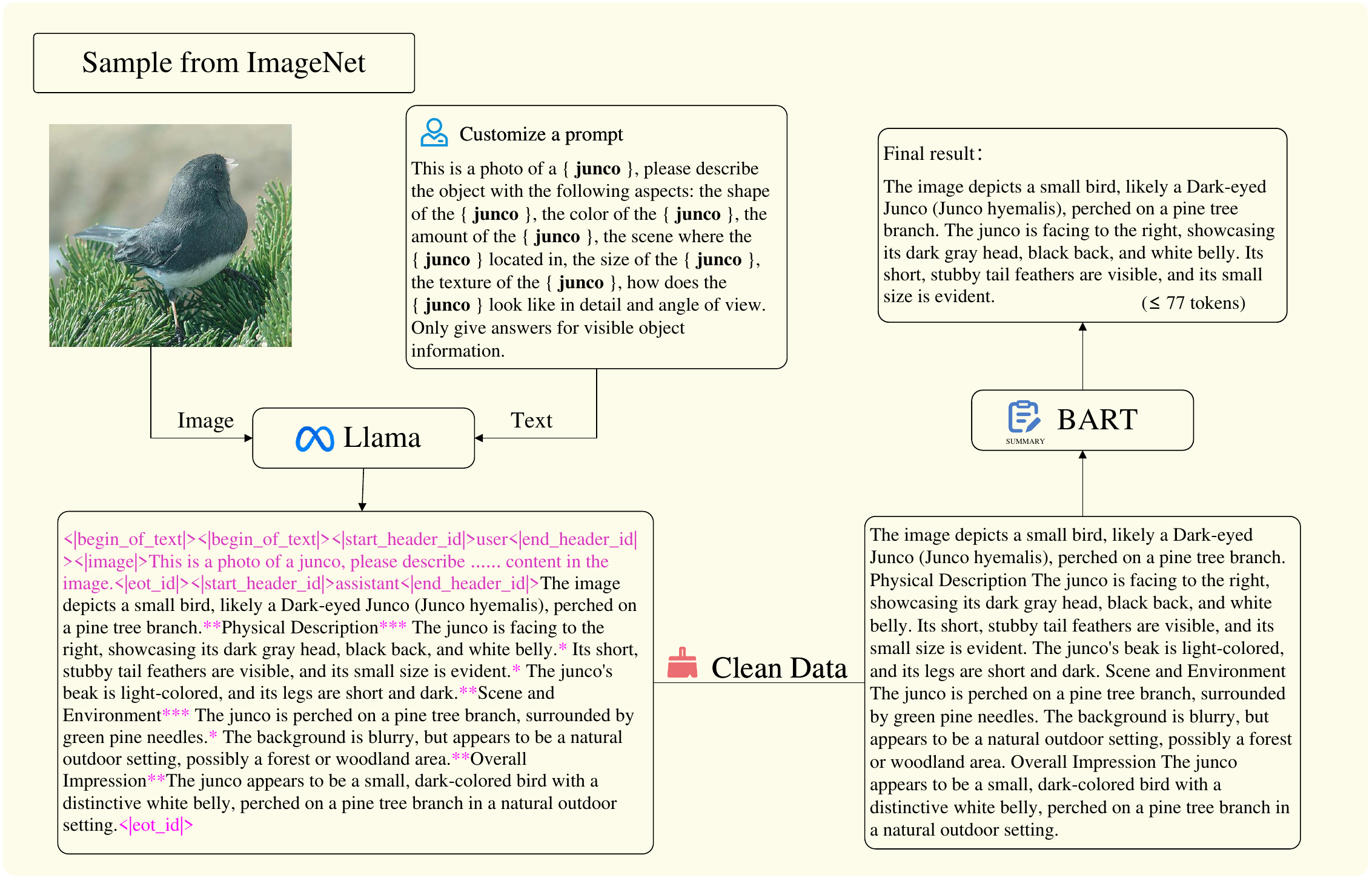}
    \caption{Pipeline of generating image description.}
    \label{fig:prompt_gen}
\end{figure}

To our knowledge, existing visual datasets generally lack corresponding image descriptions, and labelling these datasets is laborious. Therefore, we employ Llama~\cite{llama}, a multimodal large-scale model, to generate a textual description for each image. Fig.~\ref{fig:prompt_gen} illustrates the pipeline of generating image descriptions. Firstly, we customize the textual prompt for each image dataset to guide the description generating. Then, we clean the original data to reduce task-irrelevant noise (e.g. escape character, special symbol, and markdown formatting). Finally, we utilize the BART~\cite{bart} model to summarize text descriptions and compress the text length to fewer than 77 tokens, which is the maximum length of the CLIP text encoder.
\begin{figure}[ht!]
    \centering
    \includegraphics[width=0.99\linewidth]{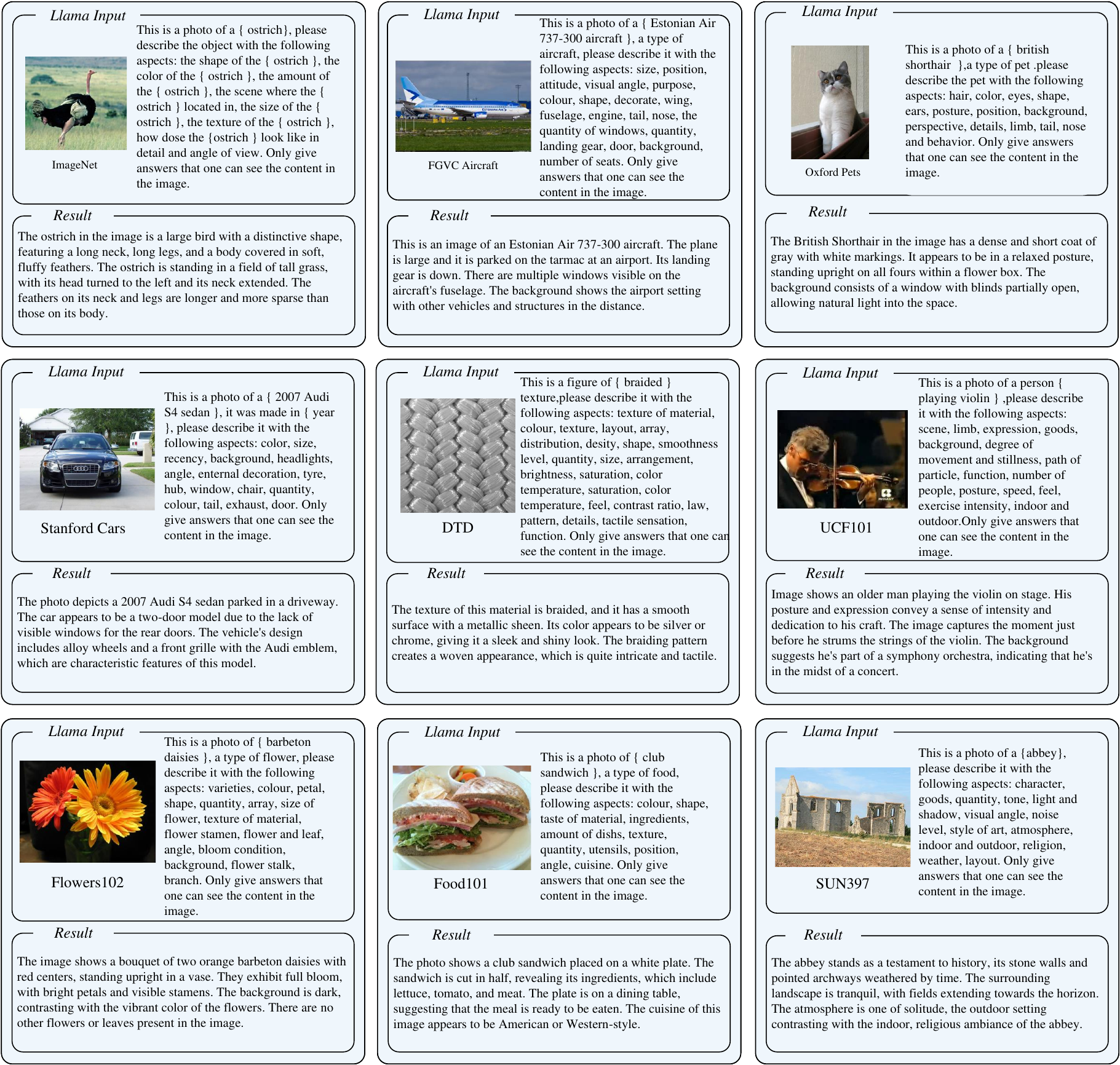}
    \caption{Examples of the image description generated by the Llama model.}
    \label{fig:prompt_example}
\end{figure}

Fig.~\ref{fig:prompt_example} shows our method for designing prompts as well as some examples of generated texts. For common vision datasets (e.g., ImageNet~\cite{imagenet}, Caltech101~\cite{caltech}), we design generalized prompts to describe the image content. We first prompt the model for the category name of the image. Then we instruct the model to describe the image's content from the following aspects: shape, color, number of objects, texture, location, and details. For fine-grained image datasets (e.g. Food101~\cite{food101}, Oxford Pets~\cite{oxfordpets}), we customize prompts to generate domain-specific image descriptions. In particular, for the Oxford Pets dataset, we prompt the model for the subclass of pets. Then, we ask the model to generate image descriptions about the pet's hair, color, eyes, shape, ears, paws, pose, and position. Fig.~\ref{fig:prompt_example} shows that the generated image descriptions are basically accurate and consistent with the image content. 

While the research on multimodel is becoming increasingly popular, large-scale image-text pair data is precious and much-needed. We supplement 11 popular image datasets (e.g., ImageNet~\cite{imagenet}, Caltech101~\cite{caltech}, and Oxford pets~\cite{oxfordpets}) by generating textual descriptions for each image, producing 1,637,795 image-text pairs in total. We name the dataset as IMD-11 and publish the dataset on the Internet for public research. 

\section{Experiment}
\label{sec:experiment}

In this section, we first describe the basic settings of the experiments and the baseline models for the comparison experiments. Next, we quantitatively and qualitatively analyze the results of the comparison experiments on 11 public datasets. Finally, we perform several ablation experiments on IDEA and T-IDEA. 

\subsection{Experiment Settings}
We select 11 popular computer vision datasets for comparison experiments, including 2 common image classification datasets (ImageNet~\cite{imagenet} and Caltech101\cite{caltech}) and 9 fine-grained image classification datasets (Food101~\cite{food101}, FGVCAircraft~\cite{fgvc}, StandCars~\cite{standcars}, UCF101~\cite{ucf101}, Flowers102~\cite{flower102}, SUN397~\cite{sun397}, EuroSAT~\cite{eurosat}, DTD~\cite{dtd} and OxfordPets~\cite{oxfordpets}). All models are trained on the training set under 1, 2, 4, 8, and 16 shots settings. For a fair comparison, the partition criteria of the training set, validation set, and test set are the same as CoOp~\cite{coop}, CLIP-Adapter~\cite{clip-adapter}, and Tip-Adapter~\cite{tip}. 

At the data pre-processing stage, we first randomly crop and scale the images with the size of $224\times 224$. Then we randomly flip and normalize the tensor of images. For the T-IDEA method, we set $50$ epochs to train the model with a bath size of $256$, and employ the stochastic gradient descent (SGD) to fine-tune the model with a learning rate of $5\times10^{-4}$. 

All experiments are conducted on a server with an AMD EPYC\textsuperscript{\texttrademark} 7642 processor, 4 NVIDIA\textsuperscript{\textregistered} GeForce RTX\textsuperscript{\texttrademark}4090 graphics cards, 256GB memory, 6TB Solid State Drive (SSD), 8TB Hard Disk Drive (HDD), and the Ubuntu 22.04.3 LTS operating system.

We compare the IDEA method and T-IDEA with five baseline models, i.e. Zero-shot CLIP~\cite{clip}, CoOp~\cite{coop}, CLIP-Adapter~\cite{clip-adapter}, Tip-Adapter\cite{tip}, and Tip-Adapter-F~\cite{tip}. All the comparison data are taken from the best result published in the original paper. To make it fair, in the comparison experiments, our method uses ResNet-50~\cite{resnet} as the visual encoder and Transformer~\cite{transformer} as the textual encoder, which is the same configuration as the five baseline models mentioned above.

\subsection{Performance Comparison and Analysis}

In this section, we conduct experiments to compare IDEA and T-IDEA with 5 baseline models on 11 publicly available image datasets.

\begin{figure*}[htp!]
    \centering
    \includegraphics[width=0.99\linewidth]{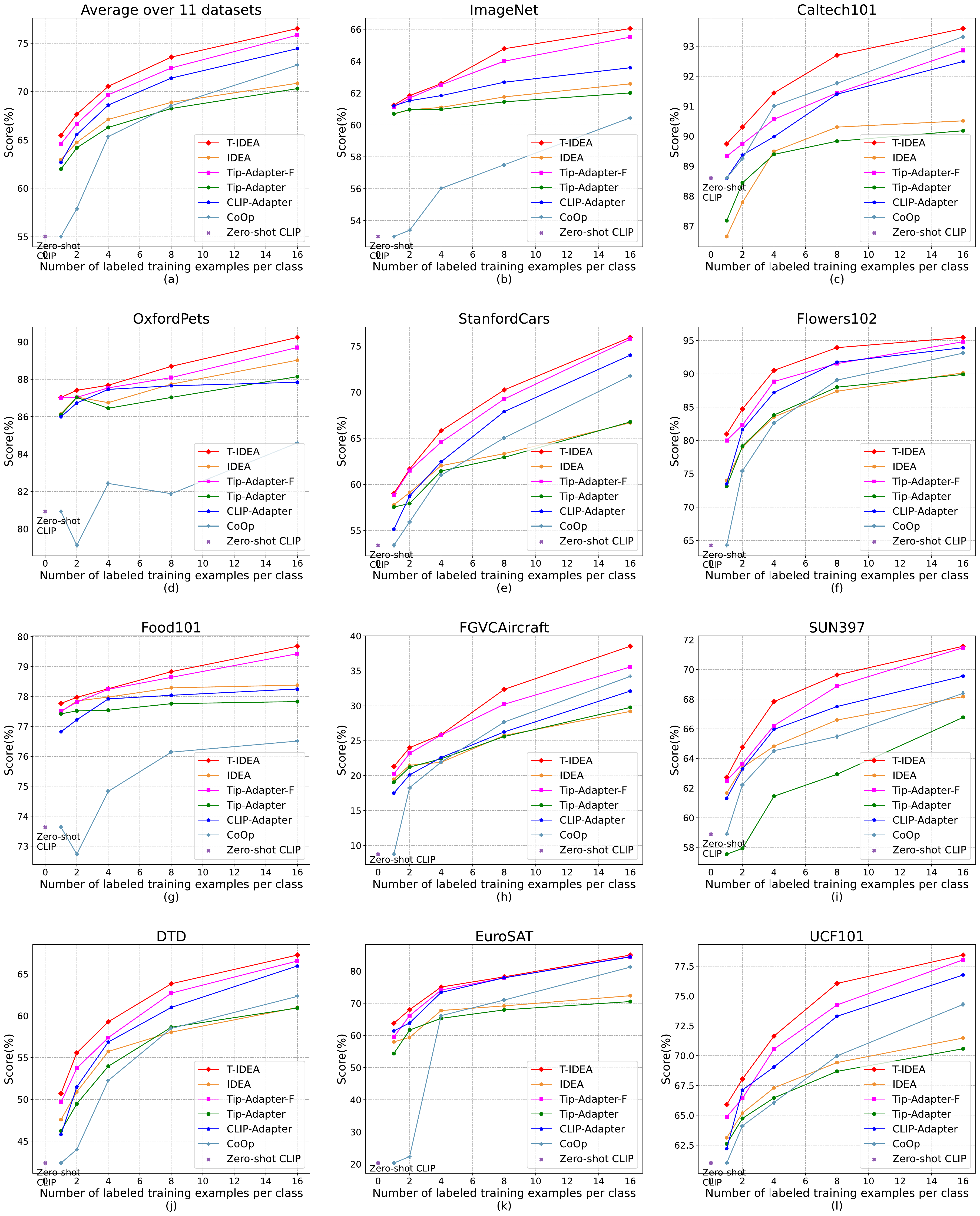}
    \caption{Results of models' performance comparison on 11 datasets. IDEA achieves good performance without extra training. T-IDEA achieves significant SOTA performance on all datasets.}
    \label{fig:exp}
\end{figure*}

Fig.~\ref{fig:exp}(a) shows the average performance of each model on the 11 image datasets. As observed,  IDEA outperforms the CoOp model, which requires additional training steps, under 1, 2, 4, and 8 shots settings. Compared to the Tip-Adapter method, which is also training-free, IDEA outperforms it by 0.63\%, 0.12\%, 0.59\%, 0.39\%, and 0.5\% under 1, 2, 4, 8, and 16 shots settings, respectively. This reveals that fusing the multimodal data (visual and textual features) in the training set is favorable to improve the model's performance. In addition, T-IDEA performs better than IDEA. As the number of shots increases, T-IDEA shows more advantages over IDEA. This phenomenon implies that it is crucial to design additional training components to fine-tune the model to better fit new features in the dataset. It is worth noticing that T-IDEA equipped with two learnable components outperforms Tip-Adapter-F under 1, 2, 4, 8, and 16 shots settings by 0.86\%, 0.99\%, 0.82\%, 1.03\%, and 0.65\%, which achieves SOTA performance.

Fig.~\ref{fig:exp}(b) and (c) indicate that both IDEA and T-IDEA methods achieve good performance in the common datasets. It is notable that on the Caltech dataset, under the 8-shot training setting, IDEA improves by 0.47\% over the Tip-Adapter method, and T-IDEA outperforms the SOTA model, Tip-Adapter-F, by 1.26\%. Fig.~\ref{fig:exp}(d-l) shows that IDEA and T-IDEA methods achieve SOTA performance in most fine-grained image classification datasets. For example, in the OxfordPets and Food101 datasets,  IDEA under the 1-shot and 2-shot settings demonstrates comparable performance with that of the SOTA model, even though the IDEA method has no extra training steps. This confirms the  advantage and superiority of IDEA especially when the category samples are limited. Meanwhile, T-IDEA achieves SOTA performance on most fine-grained image datasets. For example, on the FGVCAircraft dataset, T-IDEA outperforms Tip-Adapter-F by 2.97\% under the 16-shot setting, which is a significant boost.

In addition, we notice that our method does not perform well on some domain-specific fine-grained datasets. In Fig.~\ref{fig:exp}, we observe that, for the EuroSAT dataset, T-IDEA improves less compared to the SOTA method under 8, 16 shots settings. Given that the EuroSAT dataset is a remote sensing image classification dataset with the relatively smaller image size of $64\times 64$, it is difficult to describe the content information of the image in textual language due to the low resolution and abstract content. We infer that this may be an important reason for the limited improvement of our method on this dataset. 

\subsection{Ablation Studies}
In this section, we perform several ablation studies of IDEA and T-IDEA on the ImageNet dataset, under the 16-shot training setting, to validate the effectiveness of each component.

\subsubsection{Ablation Study of Hyperparameters}

\begin{table}[htb!]
\centering
\begin{tabular}{@{}ccccccc@{}}
\toprule
\multicolumn{7}{c}{Ablation study of Hyper-parameters}                       \\ \midrule
\multicolumn{1}{c|}{$\alpha$}    & 0     & 0.2   & 0.4   & \textbf{0.5}   & 0.8   & 1     \\ \midrule
\multicolumn{1}{c|}{Accuracy(\%)} & 59.68 & 61.36 & 62.32 & \textbf{62.58} & 62.11 & 61.63 \\ \midrule
\multicolumn{1}{c|}{$\beta$}     & 0     & 1     & 2     & 2.5   & \textbf{2.75}  & 3     \\ \midrule
\multicolumn{1}{c|}{Accuracy(\%)} & 60.34 & 61.61 & 62.28 & 62.44 & \textbf{62.58} & 62.49 \\ \midrule
\multicolumn{1}{c|}{$\theta$}    & 0.5   & 1     & 1.5   & \textbf{2}     & 3     & 3.5   \\ \midrule
\multicolumn{1}{c|}{Accuracy(\%)} & 62.05 & 62.41 & 62.49 & \textbf{62.58} & 62.43 & 62.34 \\ \bottomrule
\end{tabular}
\caption{Impact of three hyperparameters of IDEA. Bold values indicate the best results.}
\label{tab:ashp}
\end{table}

The hyperparameter $\alpha$ is designed to balance the visual similarity and the similarity of image-text pairs, as shown in Eq.~\ref{eq:sim_idea}. We set $\beta=2.75$ and $\theta=2$ and vary the value of $\alpha$ from 0 to 1. When $\alpha=0$, it means that there is only visual similarity $Sim_{I}$, and when $\alpha=1$, it means that there is only the image-text pair similarity $Sim_{T}$. Table~\ref{tab:ashp} implies that neither $Sim_{I}$ nor $Sim_{T}$ could achieve optimal performance alone. The method achieves optimal performance when $\alpha=0.5$, which indicates that visual and textual information are equally important.

The hyperparameter $\beta$ is used to  trade-off between zero-shot knowledge and few-shot knowledge, as shown in Eq.~\ref{eq:log_idea}. A larger $\beta$ indicates that more adaptation of few-shot knowledge is required. We set $\alpha=0.5,\theta=2$ and vary the value of $\beta$ from 0 to 3. $\beta=0$ means few-shot knowledge is omitted and it is the same as zero-shot CLIP. When $\beta=1$, it means that zero-shot and few-shot knowledge are of equal importance. Table~\ref{tab:ashp} shows that IDEA achieves the best performance when $\beta=2.75$, suggesting that the few-shot knowledge has a more significant weight and plays an important role in the classification results. The performance of IDEA improves by 2.24\% compared to the performance of the pure zero-shot CLIP.

The hyperparameter $\theta$ controls the sharpness of the activation function $f(x) = \text{exp}(\theta(x-1))$. With the increasing of $\theta$, the training samples which are closed to the test samples are significantly pulled apart. This operation can increase the model's ability to capture fine-grained features of images. We set $\alpha=0.5,\beta=2.75$ and vary the value of $\theta$ from 0.5 to 3.5. Table~\ref{tab:ashp} shows that IDEA achieves the best performance when $\theta=3$.

\subsubsection{Ablation Study of Trainable Components}
In this subsection, we perform ablation experiments on two learnable components (the projection layer $\mathbf{W}_{\text{proj}}$ and the semantic latent space $\mathbf{E}_{\text{bias}}$) added to T-IDEA. They are plugged and unplugged separately for a total of 4 sets of experiments.

\begin{table}[htb!]
\centering
\begin{tabular}{@{}ccc@{}}
\toprule
Projector $\mathbf{W}_{\text{proj}}$ & Semantic Latent Space $\mathbf{E}_{\text{bias}}$ & Accuracy(\%) \\ \midrule
 \XSolidBrush      &    \XSolidBrush            & 62.58      \\
 \XSolidBrush      &    \Checkmark               & 64.35       \\
 \Checkmark        &    \XSolidBrush       & 63.28      \\
 \Checkmark        &    \Checkmark                 & \textbf{66.05}    \\ \bottomrule
\end{tabular}
\caption{Ablation study on each trainable component of T-IDEA. Bold values indicate the best results.}
\label{tab:asc}
\end{table}

Table~\ref{tab:asc} shows that if we remove the projection layer $\mathbf{W}_{\text{proj}}$ from the T-IDEA, the performance decreases by 1.7\%, suggesting that the projection layer $\mathbf{W}_{\text{proj}}$ can effectively eliminate the semantic gaps between visual and textual to some degree, and achieve intermodal semantic alignment. When we remove the semantic hidden space $\mathbf{E}_{\text{bias}}$ from T-IDEA, the performance decreases by 2.77\%, indicating that the design of $\mathbf{E}_{\text{bias}}$ is able to reduce the semantic bias, leading to the performance improvement of the model. Overall, compared to the IDEA method without any trainable components, T-IDEA improves the classification metric by 3.47\%, demonstrating that combining the two components can significantly improve the model's performance.

\subsubsection{Ablation Study of Vision Backbones}

\begin{table}[h!]
\centering
\resizebox{\textwidth}{!}{
\begin{tabular}{@{}lcccc@{}}
\toprule
Model          & ResNet-50(\%) & ResNet-101(\%) & ViT-B/32(\%) & ViT-B/16(\%) \\ \midrule
Zero-Shot CLIP\cite{clip} & 60.33    & 62.53      & 63.80  & 68.73  \\
CoOp~\cite{coop} & 62.95 & 66.60 & 66.85 & 71.92 \\
CLIP-Adapter~\cite{clip-adapter} & 63.59 & 65.39 & 66.19 & 71.13 \\
Tip-Adapter~\cite{tip} & 62.03 & 64.78 & 65.61 & 70.75 \\
Tip-Adapter-F~\cite{tip} & 65.51 & 68.56 & 68.65 & 73.69 \\
\textbf{IDEA}         & 62.58     & 65.51      & 65.93  & 71.07  \\
\textbf{T-IDEA}       & \textbf{66.05}     & \textbf{68.96}   & \textbf{69.42}  & \textbf{74.54}  \\ \bottomrule
\end{tabular}
}
\caption{Performance (\%) of different models on various vision backbones. Bold values indicate the best results. ViT-B/32 denotes ViT-Base with $32 \times 32$ patch size and ViT-B/16 denotes ViT-Base with $16 \times 16$ patch size.}
\label{tab:asbb}
\end{table}

To verify the scalability of the proposed methods, we conduct further ablation experiments equipped with various backbone networks. Specifically, we replace the vision encoder in the Zero-Shot CLIP~\cite{clip}, CoOp~\cite{coop}, CIIP-Adapter~\cite{clip-adapter}, Tip-Adapter~\cite{tip}, Tip-Adapter-F~\cite{tip}, IDEA, and T-IDEA models with ResNet-50~\cite{resnet}, ResNet-101~\cite{resnet}, ViT-B/32~\cite{vit}, and ViT-B/16~\cite{vit}, respectively. From Table~\ref{tab:asbb}, we observe that under different settings of the backbone network, there is a significant performance improvement compared to the zero-shot CLIP model using only zero-shot knowledge. The performance of IDEA and T-IDEA is also improved when the parameter size of the backbone network increases. Furthermore, under different backbone settings, T-IDEA achieves SOTA performance. This indicates that our method can adapt to various backbone networks and thus demonstrates a strong generalization ability.

\section{Conclusion and Future Work}

Vision and language can semantically complement each other to enhance the ability of humans to perceive the world. Different from previous PEFT methods, we introduce a multimodal adapter to mine the multimodal information in image-text pairs, and it is fully adapted for few-shot image classification tasks. The training-free IDEA method has even outperformed the approaches that necessitate additional training steps. T-IDEA extends the IDEA method by integrating a learnable semantic alignment component and a semantic latent space component, achieving SOTA performance on 11 datasets. In addition, we design a comprehensive pipeline to generate 1.6 million image-text pairs and we publish our dataset online.

Although the performance of our methods is excellent, optimizing the text prompts could yield further enhancements. Exploring synthetic data to train models presents an intriguing area for future research. Some researchers have successfully utilized generated data from LLMs and achieved positive results~\cite{llama,llava}. In the future, we plan to investigate Chain of Thought (CoT)~\cite{cot} to generate higher-quality data from LLMs. In addition, the maximum length of input tokens in CLIP is limited to 77, constraining the amount of textual information. In the future, we will endeavor to apply IDEA and T-IDEA to Long-CLIP~\cite{longclip}. 

\bibliographystyle{elsarticle-num}



\end{document}